\documentclass[10pt,twocolumn,letterpaper]{article}

\usepackage[pagenumbers]{cvpr} 

\usepackage[utf8]{inputenc} 
\usepackage[T1]{fontenc}    
\usepackage{url}            
\usepackage{booktabs}       
\usepackage{amsfonts}       
\usepackage{nicefrac}       
\usepackage{microtype}      
\usepackage{xcolor}         

\usepackage{amsmath,amssymb,amsthm}
\usepackage{siunitx}
\usepackage[utf8]{inputenc} 
\usepackage[T1]{fontenc}    
\usepackage{url}            
\usepackage{booktabs}       
\usepackage{amsfonts}       
\usepackage{nicefrac}       
\usepackage{microtype}      
\usepackage{xcolor}         
\usepackage{enumitem} 
\usepackage{amsmath,amssymb,amsthm}
\usepackage{siunitx}
\usepackage{graphicx}
\usepackage{subcaption}

\usepackage{algorithm}
\usepackage{algorithmic}
\usepackage{multirow}
\usepackage{makecell}

\definecolor{cvprblue}{rgb}{0.21,0.49,0.74}
\usepackage[pagebackref,breaklinks,colorlinks,allcolors=cvprblue]{hyperref}

\def\confName{CVPR}
\def\confYear{2027}

\title{Improving While Deployed: Learning Proactive End-to-End Driving \\ from Imperfect Human Takeover Data}

\author{
Yangang Ren\textsuperscript{1,*},
Di Zeng\textsuperscript{2,*},
Jun Li\textsuperscript{2},
Zirui Li\textsuperscript{1},
Chengjiang Zhang\textsuperscript{1},
Xiaoqing Ye\textsuperscript{3},
Ji Tao\textsuperscript{1,$\dagger$},
Chen Lv\textsuperscript{1,$\dagger$}
\\
\vspace{0.1cm}
\textsuperscript{1}Nanyang Technological University \quad
\textsuperscript{2}Changan Automobile \quad
\textsuperscript{3}Wuhan University
\\
\textsuperscript{*}Equal contribution. \quad
\textsuperscript{$\dagger$}Corresponding authors.
\\
\texttt{taoji@changan.com.cn, lyuchen@ntu.edu.sg}
}

\begin{document}
\maketitle
\begin{abstract}
Once deployed on vehicles, end-to-end driving models continuously encounter challenging scenarios, generating valuable human takeover data that capture both model outputs and human behaviors. However, human interventions are often imperfect, as drivers may intervene too late or unnecessarily. Consequently, direct imitation can reinforce late responses or discourage otherwise appropriate maneuvers. To address this challenge, we propose an offline post-training framework that transforms imperfect takeover data into proactive supervision. We first construct diverse trajectory candidates by augmenting the plans of the deployed model and shifting recorded human corrective maneuvers earlier in time. We then train a scene-conditioned trajectory scorer to rank these candidates using expert demonstrations and unpaired preference signals from takeover logs. Using this scorer, we selectively replace pre-takeover training targets with higher-scoring candidates while retaining the original targets in normal driving scenarios. Finally, we distill the resulting targets into the driving model without changing its inference architecture. In closed-loop evaluations using real-world scenarios reconstructed with 4D Gaussian splatting, our method achieves the highest overall success rate, outperforming the deployed model, direct imitation learning and trajectory post-optimization. These results highlight a practical path toward efficient updates of deployed driving models using existing fleet logs.
\end{abstract}    
\section{Introduction}
\label{sec:intro}

End-to-end (E2E) driving models can still encounter situations that require a driver to take control after deployment on real vehicles~\cite{goff2025learning,li2026worldengine}. When a driver intervenes, the resulting log records the vehicle's
motion before takeover, the timing of the intervention, and the
driver's subsequent response. These records provide feedback for
post-training from the deployed model's own on-road
experience~\cite{yang2026human}. A simple way to use these logs is to imitate the recorded trajectories. However, drivers may take over even when the original model plan is reasonable~\cite{gershon2021driver,
zhou2025drarl}, and a necessary correction may begin only after
undesirable behavior has emerged~\cite{bi2020learning}. Direct
imitation can therefore discourage an appropriate maneuver or
reinforce a delayed response. Trajectory post-optimization can
refine the imitation target, but the result depends on predefined costs and constraints~\cite{pulver2021pilot}. The
challenge is to infer from the driver's subsequent response whether an earlier plan needed revision and, if so, how it should have changed.

One way to learn from on-road takeovers is to use driving simulators to recreate the traffic conditions that prompted them and train a model to handle similar situations~\cite{cao2023autonomous,zhou2025drarl}. For example, DRARL identifies likely causes of disengagement and evaluates the updated planner in simulated scenarios. However, the sim-to-real gap can limit how well improvements observed in these environments transfer to real vehicles~\cite{yuan2024naturalistic}. A second line of work reconstructs scenes from real driving logs at the sensor level. RAD trains E2E models through closed-loop interaction in scenes reconstructed with 3D Gaussian splatting~\cite{gao2025rad}, while SimScale combines real driving data with pseudo-expert demonstrations for synthetic variations of logged scenes~\cite{tian2026simscale}. World Engine uses fleet logs to generate safety-critical interactions for post-training and reports improvements in on-road tests~\cite{li2026worldengine}. 
While intervention signals help identify challenging scenarios, these methods primarily rely on simulated rewards or generated demonstrations, rather than directly leveraging human corrections from takeover logs.


In this paper, we propose an offline post-training framework that transforms imperfect takeover data into proactive supervision for E2E driving models. For each driving scene, we construct diverse trajectory candidates by advancing recorded human corrections and augmenting the multimodal plans of the deployed model. A scene-conditioned trajectory scorer learns to rank these candidates through expert trajectory matching and unpaired preference supervision. Unlike prior methods that directly optimize discrete action choices or action likelihoods~\cite{liu2026takead,xia2025apo}, our scorer evaluates complete trajectories from their continuous waypoint coordinates. We selectively refine pre-takeover supervision targets using its scores and distill the refined targets into the deployed model.
Our main contributions are threefold.
\begin{itemize}
    \item \textbf{Intervention-aware trajectory augmentation.}
    We shift recorded human corrections earlier by different time offsets and randomly augment the deployed model's multimodal plans. The resulting candidates offer earlier responses while retaining alternatives when corrections are unnecessary or unsuitable.

    \item \textbf{Trajectory-level preference learning.}
    We train a scene-conditioned scorer using expert trajectory matching and unpaired desirability feedback from takeover logs. It evaluates continuous, long-horizon trajectories without requiring explicit trajectory likelihoods or preference pairs involving unexecuted model plans.

    \item \textbf{Selective target refinement for offline post-training.}
    We replace pre-takeover targets only when a candidate scores sufficiently higher than the recorded trajectory and distill the refined supervision into the E2E model. Using 11,162 real-world takeover segments, our method improves the overall closed-loop success rate from 70.87\% to 81.60\%, compared with 75.30\% for direct imitation learning.
    
\end{itemize}

\section{Related Work}
\label{sec:related_works}

\textbf{Post-training for autonomous driving.}
Most E2E driving models are initially trained through imitation learning (IL) on expert demonstrations. However, optimizing trajectory similarity alone may not adequately capture safety and interaction requirements in challenging driving scenarios. Recent studies have explored reinforcement learning (RL) to further improve these models. WorldRFT \cite{yang2026worldrft} employs collision-aware trajectory rewards and group relative policy optimization to enhance planning safety. DiffusionDriveV2 \cite{zou2025diffusiondrivev2} introduces trajectory exploration with intra- and inter-anchor optimization to improve trajectory diversity and quality. PaIR-Drive \cite{lian2026pairdrive} integrates IL and RL through parallel optimization to mitigate policy drift.
Another line of research explores closed-loop training through world models and reconstructed environments. Raw2Drive \cite{yang2025raw2drive} aligns privileged and raw-sensor world models to enable RL-based training. RAD \cite{gao2025rad} employs 3DGS-based closed-loop RL in reconstructed driving scenes, while RAD-2 \cite{gao2026rad2} introduces a generator-discriminator framework with temporally consistent policy optimization and efficient BEV-based simulation. World Engine \cite{li2026worldengine} mines safety-critical scenarios from real-world driving logs and synthesizes diverse traffic interactions for post-training. 

\noindent\textbf{Learning from human interventions.}
Human feedback provides additional supervision for improving driving models beyond fixed expert demonstrations. DAgger~\cite{ross2011reduction} queries expert action labels for states visited by the model, while HG-DAgger~\cite{kelly2019hgdagger} collects corrective demonstrations through human takeovers. In robotic manipulation, human interventions have been used to reweight imitation data~\cite{liu2025robot}, derive rewards~\cite{luo2024rlif}, and learn action preferences~\cite{xia2025apo}. In autonomous driving, DICL~\cite{cao2023autonomous} constructs simulated training experiences from real-world intervention cases, while DRARL~\cite{zhou2025drarl} identifies takeover reasons and generates similar scenarios for model improvement. CAT~\cite{zhang2023cat} further augments log-replayed driving scenarios through adversarial traffic generation for closed-loop training. Recent E2E studies also exploit simulated takeover data. TakeAD~\cite{liu2026takead} and CoReVLA~\cite{fang2025corevla} collect takeover demonstrations from privileged experts and human drivers in simulated environments, respectively. TakeVLA~\cite{gao2026learning} incorporates pre-takeover language supervision and reinforcement finetuning through scenario dreaming. However, real-world interventions may occur without planning errors~\cite{zhou2025drarl} or after an appropriate corrective maneuver should have begun~\cite{bi2020learning}. Our method directly exploits these imperfect corrections to refine trajectory-level supervision offline, enabling efficient updates to deployed E2E models.

\noindent\textbf{Preference learning for model refinement.}
Preference learning enables model improvement by distinguishing desirable from undesirable behaviors. Representative approaches, including DPO \cite{rafailov2023direct} and SimPO \cite{meng2024simpo}, optimize models using chosen-rejected response pairs obtained from human judgments or reward-based rankings. In contrast, KTO \cite{ethayarajh2024kto} learns from unpaired binary desirability feedback without explicit preference pairs. In autonomous driving, TakeAD \cite{liu2026takead} leverages takeover preferences to optimize distributions over discrete trajectory modes and control actions. DriveDPO \cite{shang2025drivedpo} constructs trajectory preference pairs using imitation similarity and simulation-based safety scores, while EMPlan \cite{chen2026emplan} employs reward-guided unpaired preference optimization for multimodal planning. In robotic manipulation, APO \cite{xia2025apo} exploits binary desirability feedback from human interventions to refine vision-language models. These approaches primarily apply preference supervision directly to model optimization, often relying on action or trajectory likelihoods.
Instead, we learn a scene-conditioned scorer from imperfect real-world takeover data to evaluate candidate coordinates directly and refine trajectory supervision for distillation without using the driving model's trajectory likelihoods.

\section{Methodology}
\label{sec:method}

\subsection{Overview}

\begin{figure*}[!htbp]
    \centering
    \includegraphics[width=\linewidth]{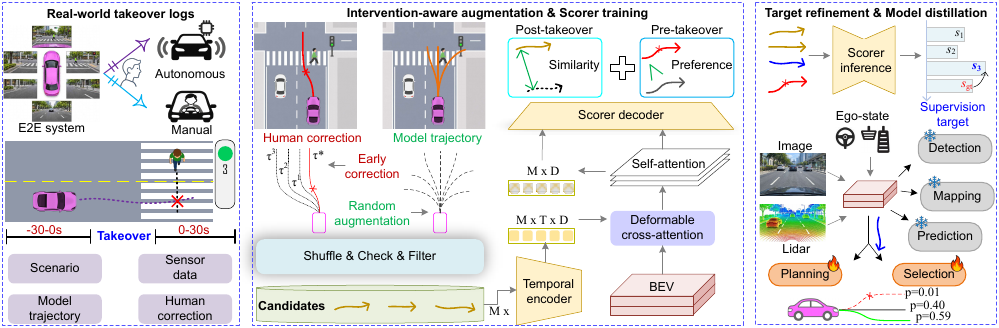}
    \caption{Overview of the proposed offline post-training framework. \textbf{Left}: Real-world takeover logs capture the model's planning results and subsequent human corrections. \textbf{Middle}: Diverse trajectory candidates are constructed through temporal advancement of human corrections and random augmentation of model trajectories. A scene-conditioned trajectory scorer is trained using expert trajectory matching and unpaired preference supervision. \textbf{Right}: The learned scorer selectively refines pre-takeover supervision targets, which are distilled into the E2E driving model without modifying its inference architecture.}
    \label{fig:framework}
\end{figure*}

We consider an E2E driving model trained on large-scale expert driving data. The model takes seven synchronized camera views and LiDAR point clouds as inputs. Image features are extracted using a ResNet-50 backbone \cite{he2016deep} with a feature pyramid network \cite{lin2017feature}, while point clouds are encoded using a PointPillars-based encoder \cite{lang2019pointpillars}. Features from both modalities are fused into a unified bird's-eye-view (BEV) representation \cite{liu2022bevfusion}. As shown in Figure~\ref{fig:framework}, multiple task-specific heads perform object detection \cite{zhang2023dino}, motion prediction, online vectorized mapping \cite{liao2025maptrv2}, and trajectory planning \cite{liu2025hybrid}. Conditioned on the BEV features and ego states, the planning head generates $N=32$ multimodal trajectories over a 6-second horizon, each containing $T=60$ waypoints. Each waypoint is parameterized by the mean and variance of its 2D position, yielding an output tensor of $N\times T\times4$. A selection head assigns confidence scores to the trajectories, selecting the highest-ranked candidate for vehicle control. The model replans at 10 Hz, so only the initial portion of each selected 6-second trajectory is executed before the next planning cycle. The entire E2E model comprises approximately 140 million parameters.

\begin{figure}[!hbtp]
    \centering
    \begin{subfigure}{0.48\columnwidth}
        \centering
        \includegraphics[width=\linewidth]{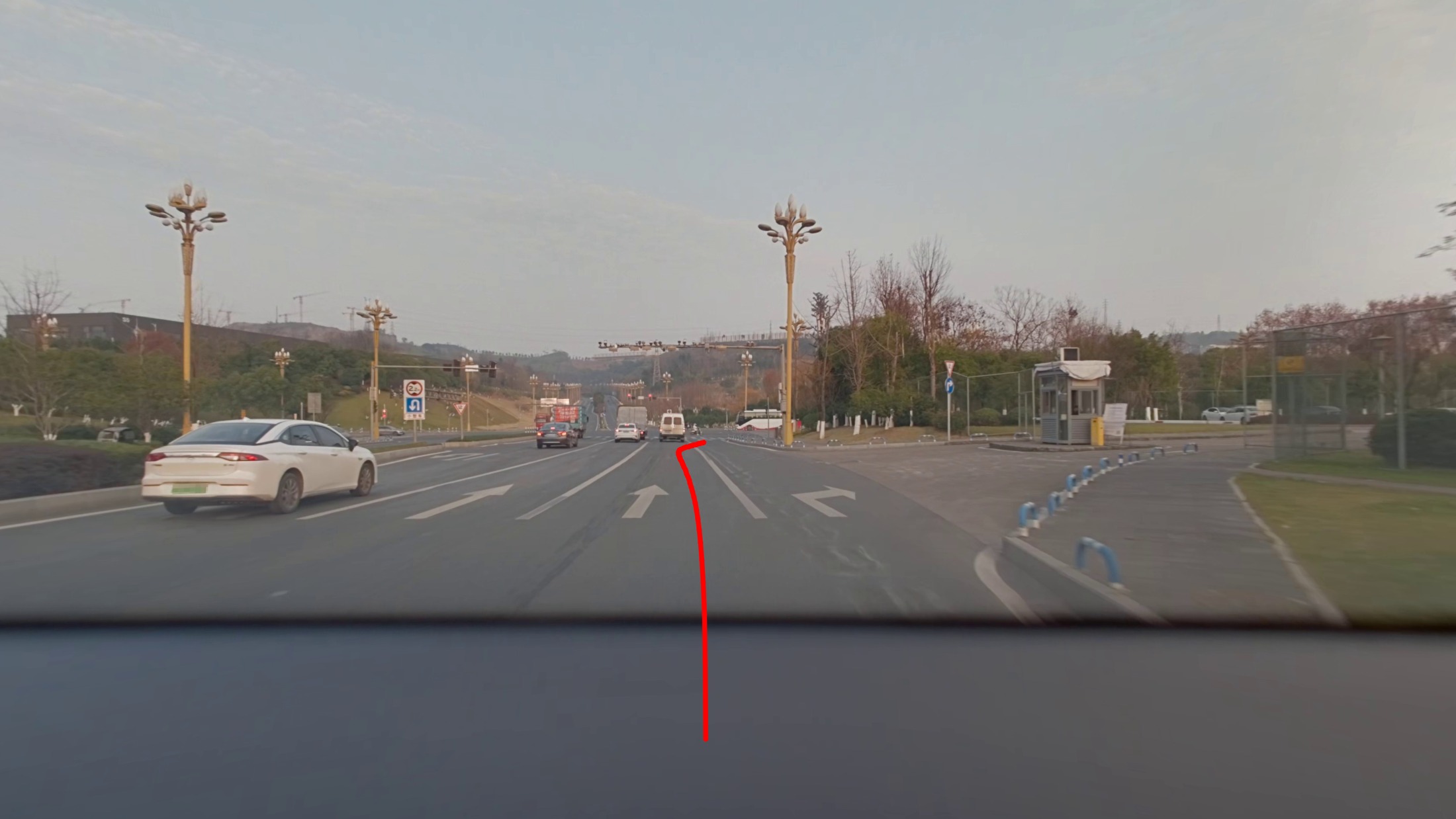}
        \par\vspace{0pt}
        \includegraphics[width=\linewidth]{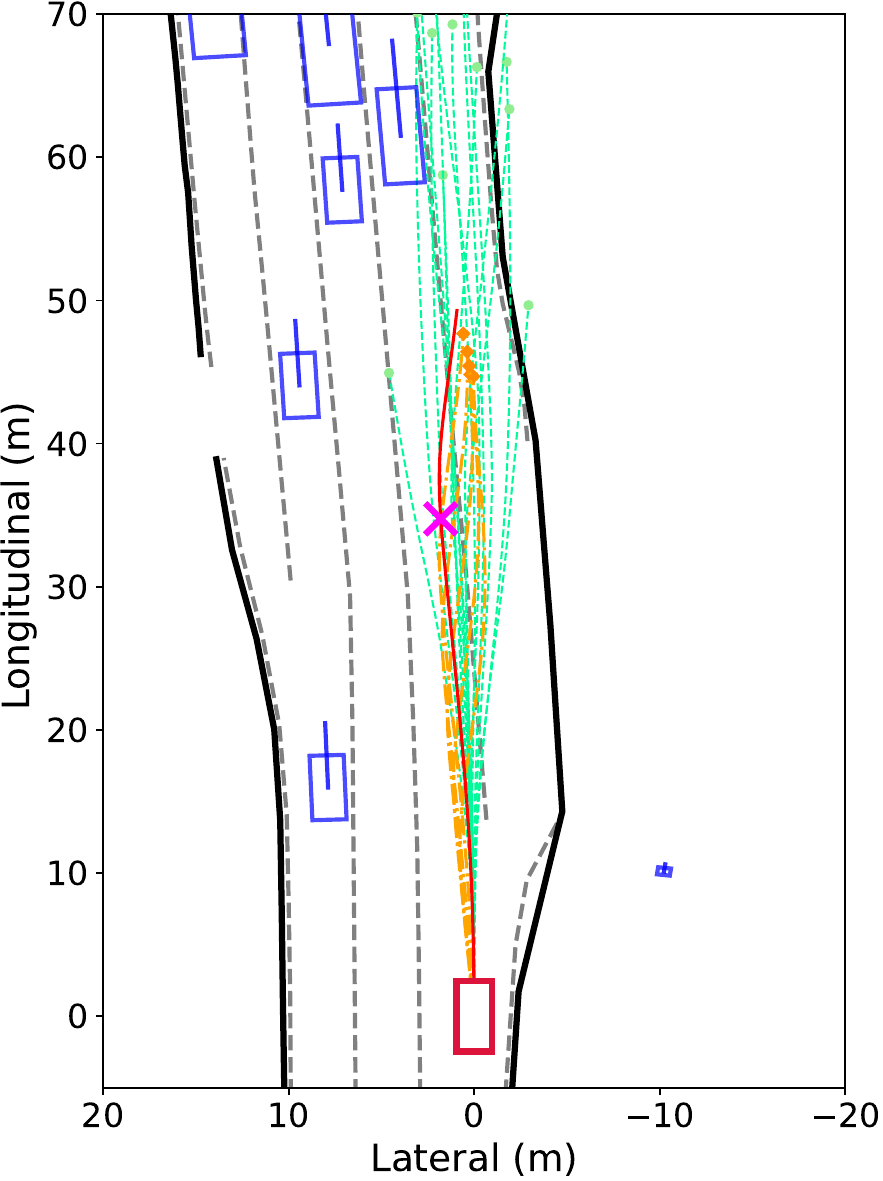}
        \caption{}
        \label{fig:takeover_a}
    \end{subfigure}%
    \hfill
    \begin{subfigure}{0.48\columnwidth}
        \centering
        \includegraphics[width=\linewidth]{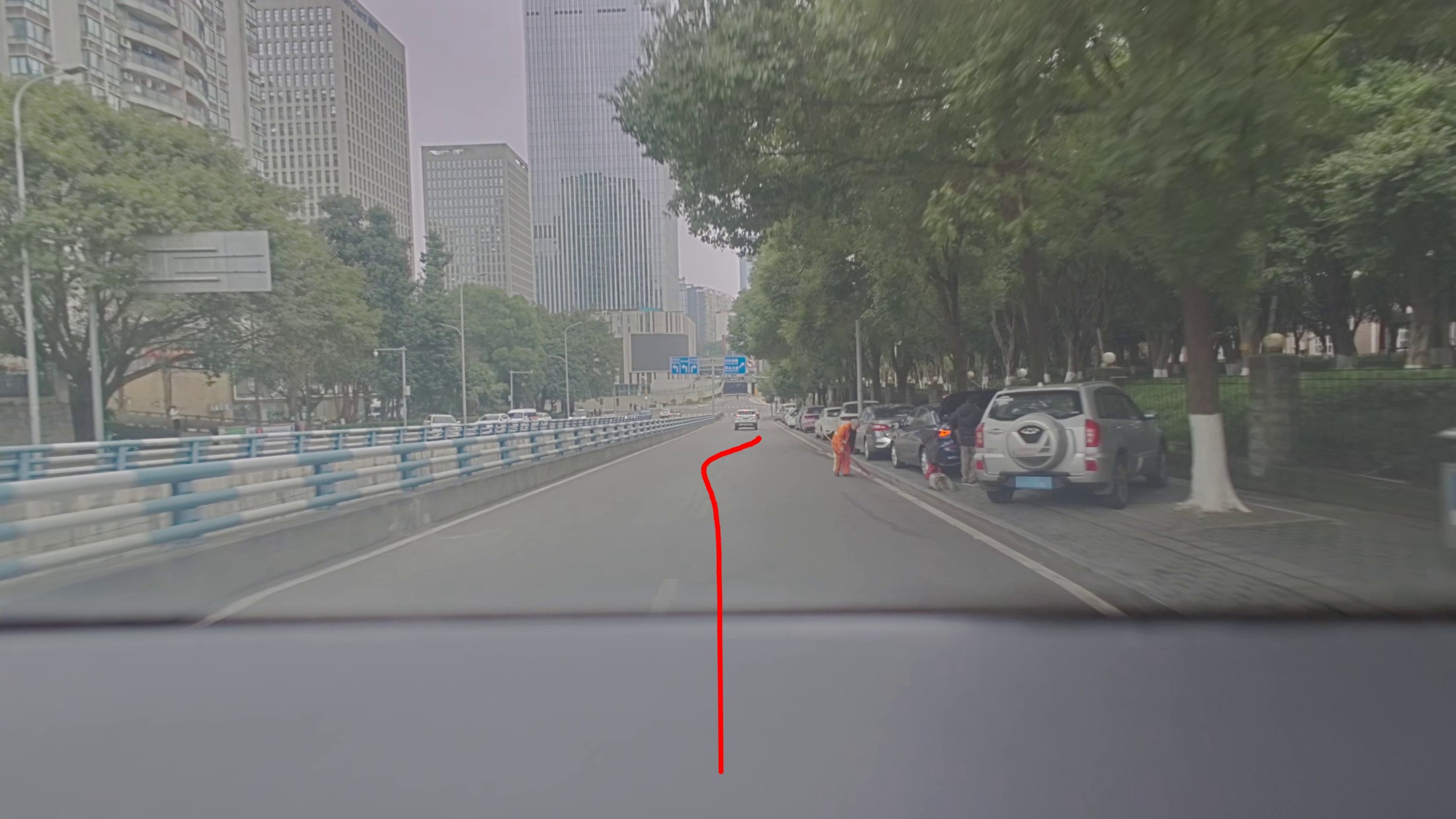}
        \par\vspace{0pt}
        \includegraphics[width=\linewidth]{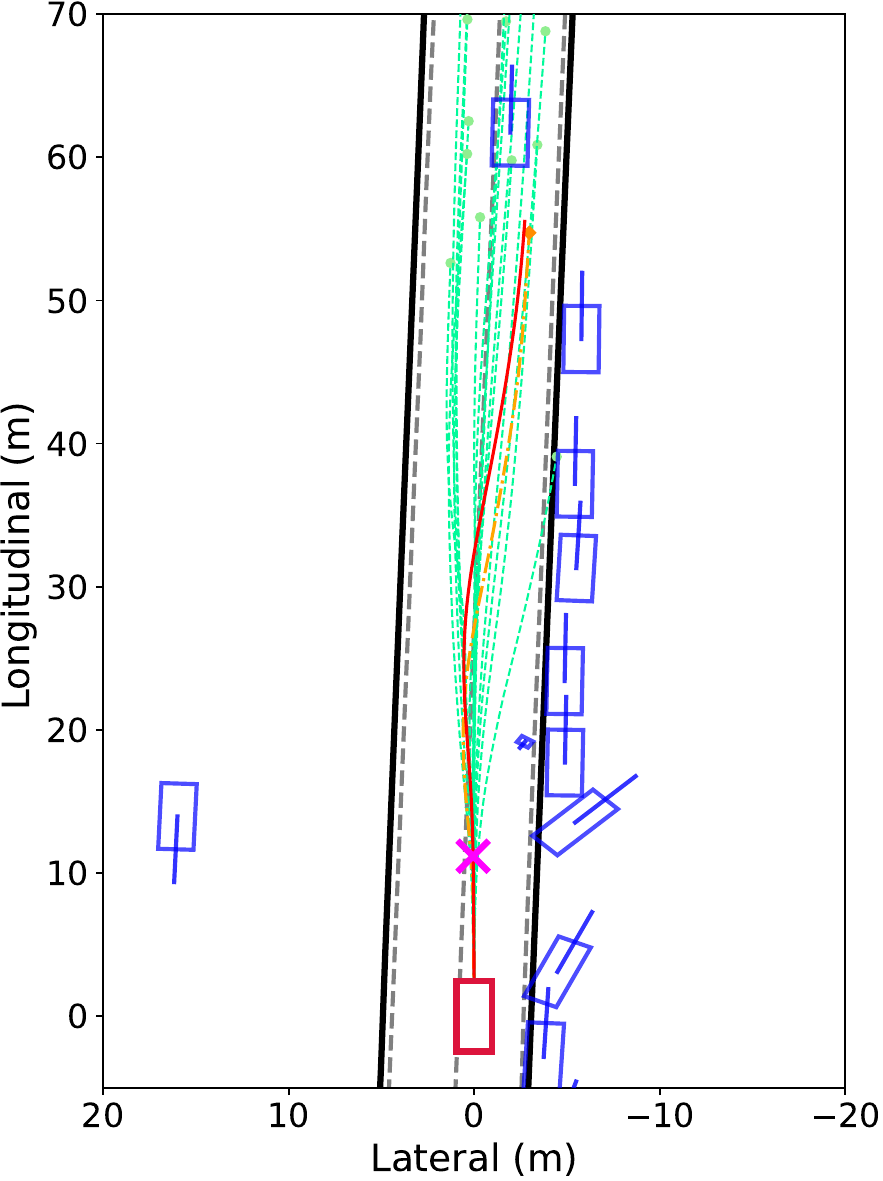}
        \caption{}
        \label{fig:takeover_b}
    \end{subfigure}


    \begin{subfigure}{0.48\columnwidth}
        \centering
        \includegraphics[width=\linewidth]{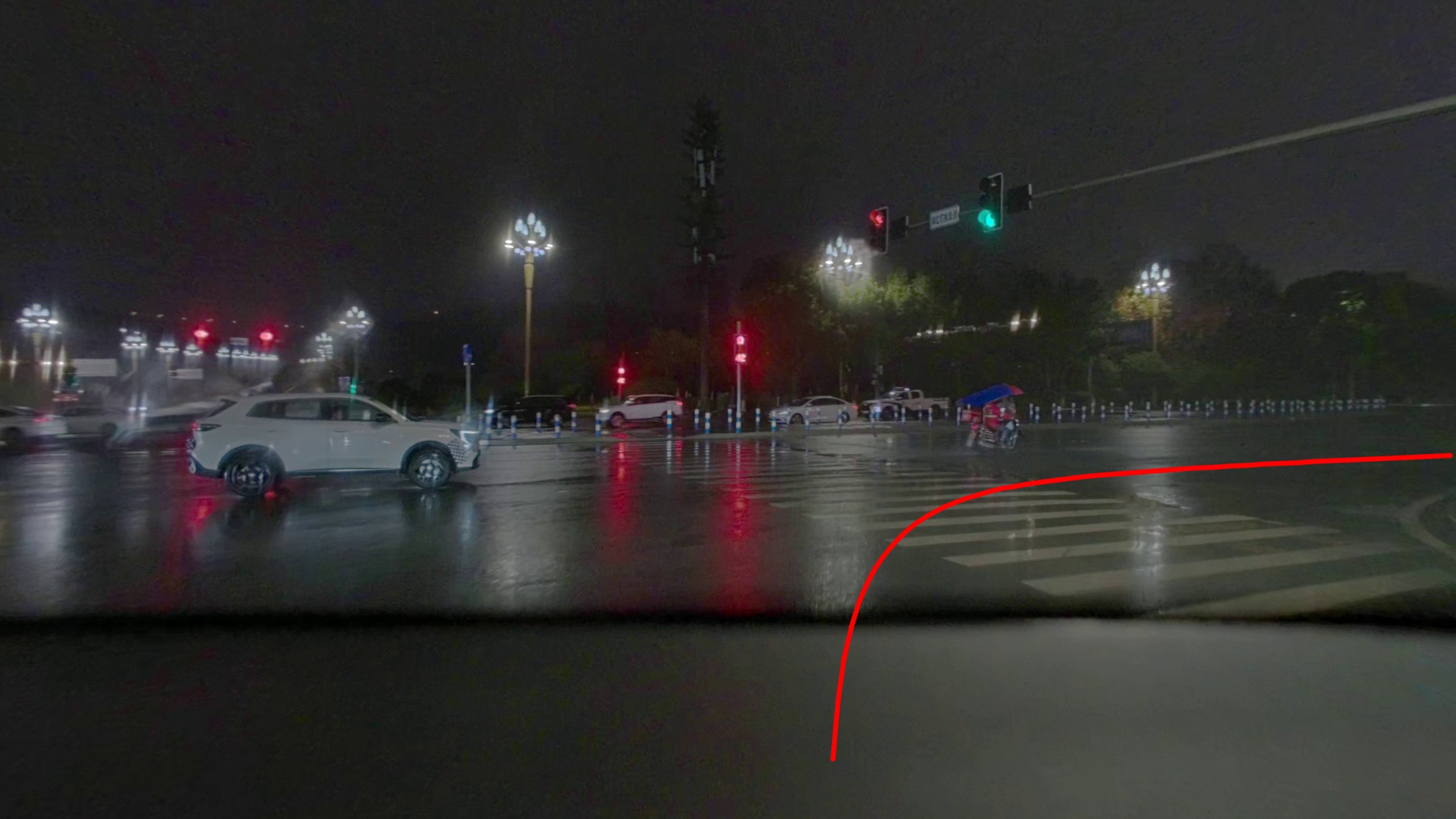}
        \includegraphics[width=\linewidth]{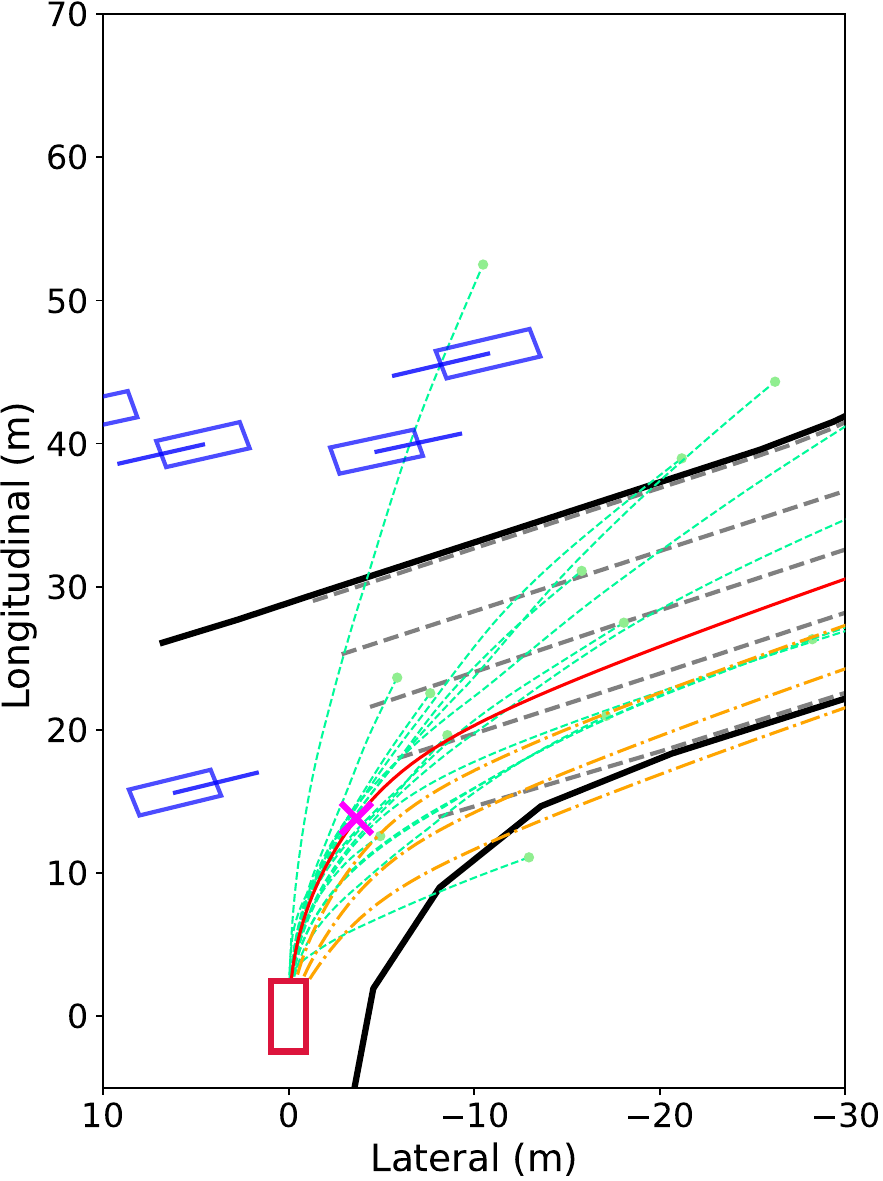}
        \caption{}
        \label{fig:takeover_c}
    \end{subfigure}%
    \hfill
    \begin{subfigure}{0.48\columnwidth}
        \centering
        \includegraphics[width=\linewidth]{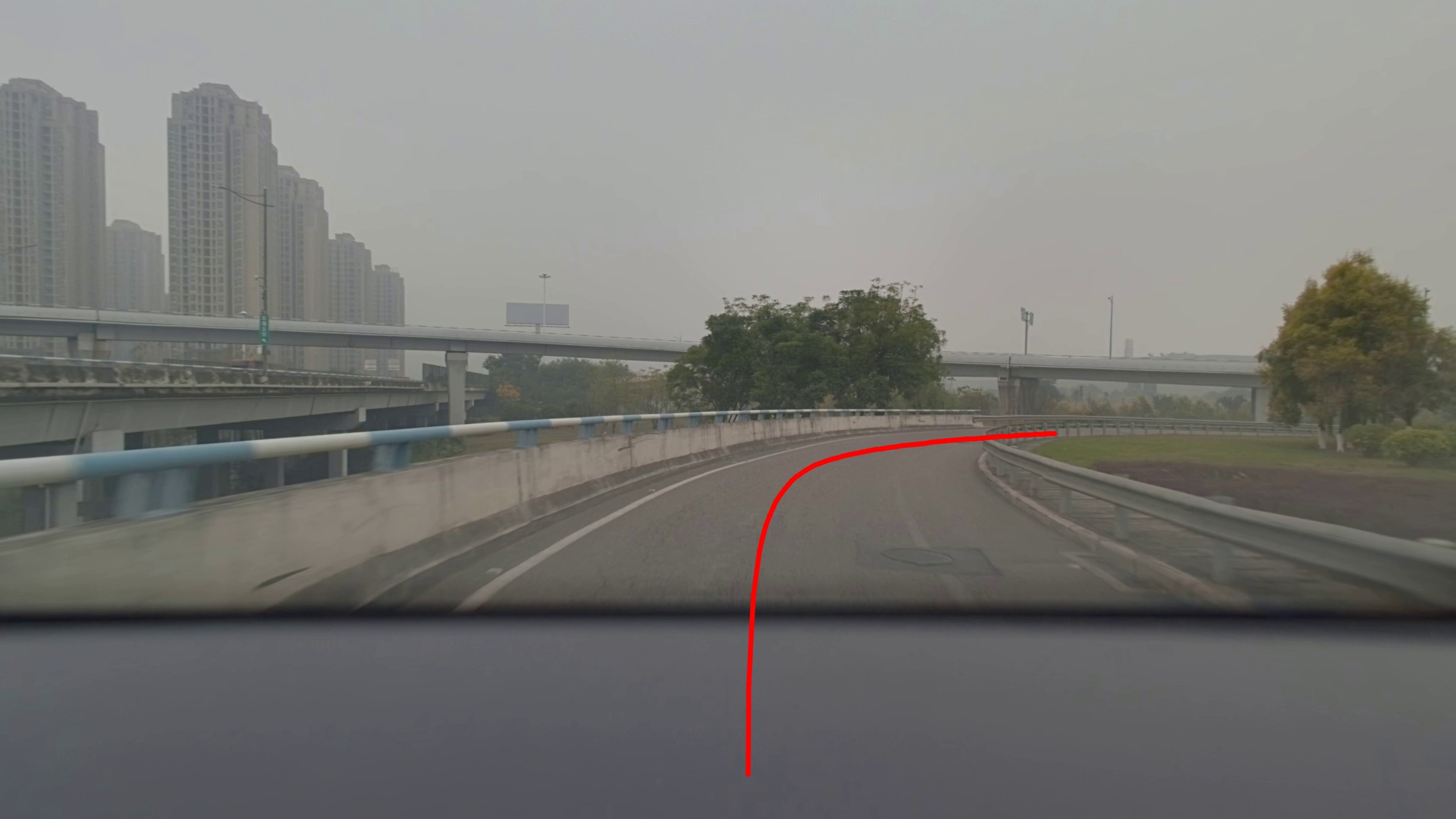}
        \includegraphics[width=\linewidth]{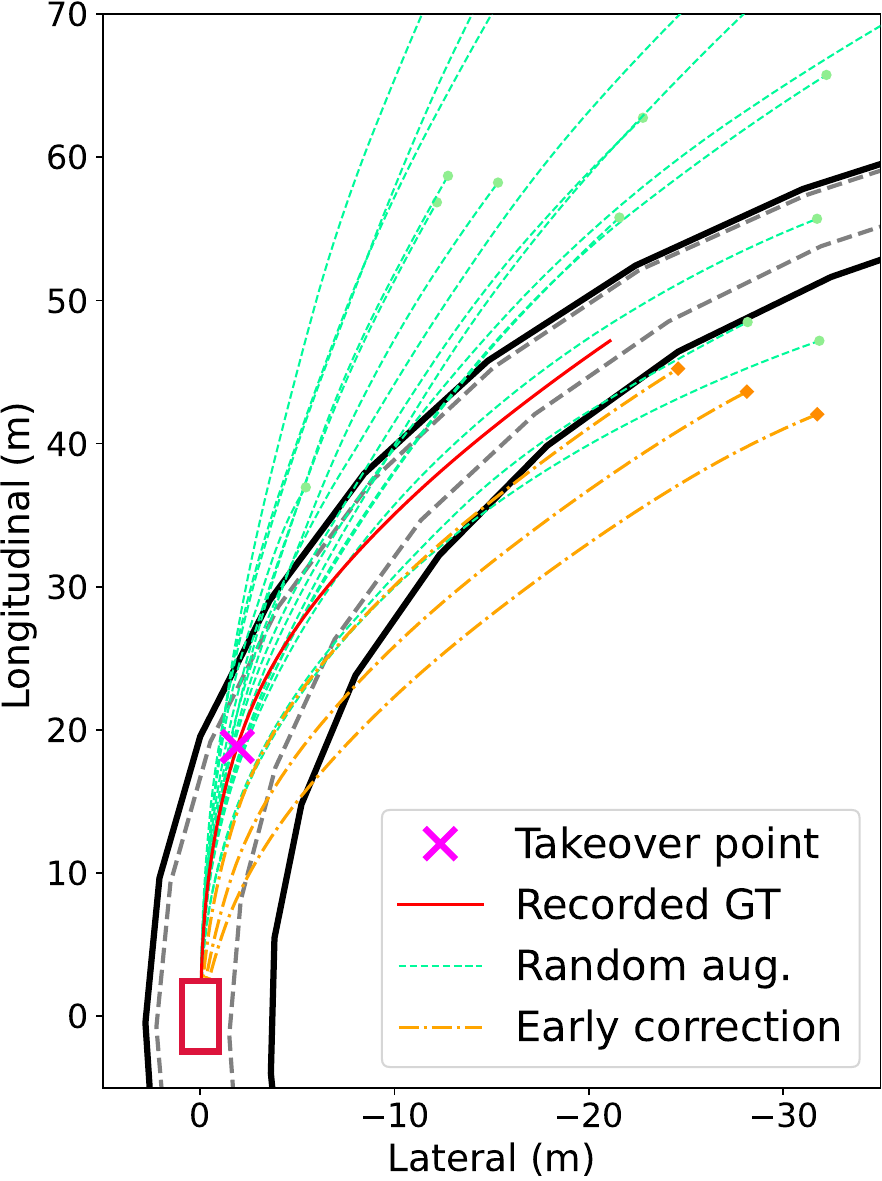}
        \caption{}
        \label{fig:takeover_d}
    \end{subfigure}

    \caption{Results of trajectory augmentation in four representative real-world takeover scenarios. Each case shows the onboard camera image (top) and augmented trajectories in BEV (bottom).
    (a) Failure to follow right-turn navigation.
    (b) Insufficient avoidance of a sanitation worker.
    (c) Potential collision during a nighttime right turn.
    (d) Curb collision risk on a sharply curved ramp.}
    \label{fig:traj_aug}
\end{figure}

The trained E2E model is deployed as a TensorRT engine on the onboard computing platforms of production vehicles. Real-world driving data are collected from diverse urban environments across Chongqing, Beijing and Shanghai in China. The vehicles feature different sensor configurations, including single- and dual-LiDAR setups, with sensor inputs adapted to a unified representation for training and inference. The E2E driving system operates under human supervision, requiring drivers or safety testers to intervene when necessary. For each takeover event, the vehicle automatically records a 60-second driving segment spanning 30 seconds before and after the intervention. The logs include synchronized sensor observations, scenario information, model outputs, and subsequent human corrections. The raw takeover logs cover diverse driving issues, primarily involving navigation (35.8\%), lane changes (33.5\%), and obstacle avoidance (22.3\%), with the remainder related to braking, traffic signals, and other issues. Figure~\ref{fig:traj_aug} illustrates four representative takeover scenarios: failure to follow right-turn navigation (Figure~\ref{fig:takeover_a}), insufficient avoidance of a sanitation worker entering the driving path (Figure~\ref{fig:takeover_b}), potential collision with oncoming traffic during a nighttime right turn (Figure~\ref{fig:takeover_c}), and curb collision risk on a sharply curved ramp (Figure~\ref{fig:takeover_d}). Using an established industrial data-processing pipeline, we filter out samples with missing sensor data or discontinuous driving trajectories, yielding 11,162 takeover segments and 4,011,830 valid frames for post-training.

\subsection{Intervention-aware trajectory augmentation}

Human takeovers enable vehicles to safely navigate
challenging real-world scenarios where deployed models
exhibit undesirable behaviors. A straightforward approach
is to imitate these recorded trajectories.
However, human corrections are subject to imperfect
intervention decisions \cite{xue2025policy} and temporal
delays \cite{liu2025robot,zhu2026improving}. Consequently, direct imitation may reinforce delayed or suboptimal reactions \cite{luo2024rlif} and introduce supervision that conflicts with the acquired capabilities. At time step $t$, we denote the logged 6-second driving trajectory as $\boldsymbol{\tau}_t^* =\{\mathbf{p}_t(k\delta)\}_{k=1}^{T}$, where $\delta=0.1$ s and $\mathbf{p}_t(s)$ represents the recorded ego position at time $t+s$, expressed in the local coordinate frame at $t$. The driving logs also provide extended future trajectories beyond the 6-second planning horizon. For a takeover at $t_o$ with $t<t_o<t+T\delta$, $\boldsymbol{\tau}_t^*$ combines autonomous driving behavior before the intervention with subsequent human corrections, potentially providing suboptimal supervision at time $t$.

To compensate for intervention delays, we generate $N_e$ augmented trajectories by advancing the recorded human corrections with different time offsets $\Delta_j=0.5j$ seconds, where $j=1,\ldots,N_e$. Each augmented trajectory is constructed as
\begin{equation}
\boldsymbol{\tau}_{t,j}
= \left\{
\mathbf{p}_t(k\delta+\Delta_j)
- \mathbf{p}_t(\Delta_j)
\right\}_{k=1}^{T},
\label{eq:1}
\end{equation}
where the translation aligns the shifted trajectory with the current ego origin. This advances the onset of human correction from $t_o$ to $t_o-\Delta_j$ while preserving the recorded motion profile. We fill the tail of each shifted trajectory with subsequent logged ego positions, preserving the full 6-second planning horizon without extrapolation. To limit initial-state mismatch, we retain shifted trajectories only if their initial speed and heading differ from the current values by at most $5\, \mathrm{m/s}$ and $30^\circ$, respectively. Otherwise, they are replaced with random perturbations of the recorded trajectory to maintain $N_e$ candidates.

In addition to early human corrections, we augment the multimodal
planned trajectories of the E2E model to further expand the
candidate space. This provides alternative driving behaviors
when human takeovers are unnecessary or advanced corrections
are unsuitable. Given $N=32$ planned trajectories
$\{\boldsymbol{\tau}_t^i\}_{i=0}^{N-1}$, we apply scale-adaptive
multiplicative Gaussian noise \cite{zou2025diffusiondrivev2}
to generate $N_{\mathrm{aug}}$ variants for each trajectory:
\begin{equation}
    \boldsymbol{\tau}_{t,j}^{i}
    = (1+\boldsymbol{\epsilon}_{i,j})
    \odot \boldsymbol{\tau}_t^i,
    \label{eq:trajectory_augmentation}
\end{equation}
where $j=1,\ldots,N_{\mathrm{aug}}$ and
$\boldsymbol{\epsilon}_{i,j}
=(\epsilon_{\mathrm{long}}^{i,j},
\epsilon_{\mathrm{lat}}^{i,j})$
denotes independently sampled perturbations along the longitudinal and lateral directions.
Each perturbation is applied only to the predicted position means and shared across all waypoints of the trajectory to preserve temporal consistency.
Combining the original model trajectories, their augmented variants, the recorded GT, and the $N_e$ early correction variants yields a candidate set $\mathcal{C}_t$ of size $M=|\mathcal{C}_t|=N(N_{\mathrm{aug}}+1)+1+N_e$ for scorer training.

Figure~\ref{fig:traj_aug} illustrates the complementary effects of the two augmentation strategies across four representative takeover scenarios. Advancing human corrections enables earlier maneuvers, such as a rightward lane change before turning in Figure~\ref{fig:takeover_a} and an earlier leftward evasive maneuver to avoid a pedestrian in Figure~\ref{fig:takeover_b}. However, some shifted trajectories extend beyond the drivable area in Figure~\ref{fig:takeover_c} and Figure~\ref{fig:takeover_d}, highlighting the need for alternative candidates. Model trajectories and their random perturbations provide additional alternatives when shifted corrections are unsuitable. Throughout augmentation, the recorded road geometry and surrounding obstacle trajectories remain unchanged, as the augmented trajectories serve as candidate proposals for subsequent scoring rather than physically consistent rollouts.

\subsection{Preference learning for trajectory scoring}
\label{sec:preference_scorer}

To evaluate the candidate set $\mathcal{C}_t$, we introduce a Transformer-based trajectory scorer $f_{\phi}$ that takes the BEV features $z_t$ and $\mathcal{C}_t$ as inputs, as illustrated in Figure~\ref{fig:framework}.
Each candidate is processed by a temporal encoder, followed by deformable cross-attention with BEV features and self-attention among candidate embeddings.
During training, the recorded ground truth (GT) is included in $\mathcal{C}_t$ for supervision, with a self-attention mask preventing other candidates from attending to it.
A scoring head assigns a scalar desirability score $s_t^i=[f_{\phi}(z_t,\mathcal{C}_t)]_i$ to each candidate. For brevity, we write $s(\boldsymbol{\tau}_t^{i})=s_t^i$, with the scene context implicit.
To account for imperfect takeover trajectories, we combine cross-entropy (CE) supervision for trajectory matching with unpaired preference (UP) learning.
The CE classification target is the candidate with the minimum average displacement error (ADE) to the GT:
\begin{equation}
    i^* = \arg\min_{i\in\mathcal{I}_t}
    \mathrm{ADE}(\boldsymbol{\tau}_t^{i},
    \boldsymbol{\tau}_t^*).
    \label{eq:scorer_ce_target}
\end{equation}
The CE loss is defined as
\begin{equation}
    \mathcal{L}_{\mathrm{CE}} =
    -\mathbb{E}_{t\sim\mathcal{D}_{\mathrm{CE}}}
    \left[
    \lambda_t
    \log\frac{\exp(s_t^{i^*})}
    {\sum_{i\in\mathcal{I}_t}\exp(s_t^i)}
    \right],
    \label{eq:scorer_ce}
\end{equation}
where $\mathcal{D}_{\mathrm{CE}}$ denotes all training samples, and $\mathcal{I}_t$ contains the eligible candidate indices, excluding the recorded GT to prevent trivial selection. The sample-specific weight is $\lambda_t=0.1$ for negative samples and $\lambda_t=1$ otherwise.

For pre-takeover trajectories, trajectory-matching supervision alone may not reliably reflect driving desirability. Previous methods such as TakeAD \cite{liu2026takead} construct pairwise preferences over discrete trajectory modes and control actions. However, such comparisons are less reliable for long-horizon trajectories, as model-generated plans are not fully executed. Inspired by APO \cite{xia2025apo}, we instead adopt unpaired preference (UP) learning to suppress intervention-prone trajectories without constructing explicit preference pairs. Specifically, samples whose future 6-second trajectory contains a takeover event, i.e., $0<t_o-t\leq6\,\mathrm{s}$, are treated as negative samples, while normal autonomous driving and post-takeover human driving provide supervision through CE.
Let $\mathcal{D}^{-}$ denote the negative samples, we define the unpaired preference loss as
\begin{equation}
\begin{aligned}
\tilde{s}_t
&= \log\left(
\frac{1}{|\mathcal{I}_t|}
\sum_{i\in\mathcal{I}_t}
\exp\big(s(\boldsymbol{\tau}_t^{i})\big)
\right),\\
\mathcal{L}_{\mathrm{UP}}
&= \mathbb{E}_{t\sim\mathcal{D}^{-}}
\left[
\operatorname{softplus}
\left(s(\boldsymbol{\tau}_t^*)-\tilde{s}_t\right)
\right],
\end{aligned}
\label{eq:scorer_preference}
\end{equation}
where $\tilde{s}_t$ denotes the log-mean-exp score of alternative trajectories. This objective suppresses the relative score of intervention-prone GT without assigning explicit preference labels to unexecuted model plans.

The overall scorer training objective combines the CE and UP losses:
\begin{equation}
    \mathcal{L}_{\mathrm{scorer}}
    = \mathcal{L}_{\mathrm{CE}}
    + \mathcal{L}_{\mathrm{UP}}.
    \label{eq:scorer_total}
\end{equation}
The CE loss encourages expert-aligned trajectory selection, while the UP loss calibrates trajectory desirability based on takeover feedback. The BEV feature extractor remains frozen during scorer training.

\subsection{Target refinement and model distillation}
\label{sec:policy_distillation}

With the trained trajectory scorer $f_{\phi}$, we refine the supervision targets for post-training the E2E driving model. Specifically, we load and freeze a copy of the deployed model to generate multimodal trajectory proposals and construct the augmented candidate set $\mathcal{C}_t$ using the intervention-aware augmentation strategy. The scorer selects the highest-scoring trajectory $\boldsymbol{\tau}_t^{\bar{i}}$, with $\bar{i}=\arg\max_i s(\boldsymbol{\tau}_t^i)$, as a potential replacement for the recorded trajectory $\boldsymbol{\tau}_t^*$.

To avoid unnecessary modifications to reasonable driving behaviors, we restrict target refinement to the 6-second window preceding each takeover. Moreover, the selected candidate's score must exceed that of the recorded trajectory by more than 0.1. Formally, the refined target is defined as
\begin{equation}
\bar{\boldsymbol{\tau}}_t^* =
\begin{cases}
\boldsymbol{\tau}_t^{\bar{i}},
& \begin{aligned}
  & 0<t_o-t \le 6 \mathrm{s}, s_t^{\bar{i}}-s_t^{\mathrm{gt}}>0.1,
  \end{aligned}\\
\boldsymbol{\tau}_t^*, & \text{otherwise},
\end{cases}
\label{eq:target_refinement}
\end{equation}
where $\boldsymbol{\tau}_t^{\bar{i}}$ denotes the highest-scoring candidate in $\mathcal{C}_t$, and $s_t^{\mathrm{gt}}$ is the score of the recorded trajectory. This selective refinement focuses supervision updates on intervention-prone states while preserving the original targets elsewhere.

Finally, we perform supervised post-training using the refined targets $\bar{\boldsymbol{\tau}}_t^*$. We freeze all other components of the E2E model and finetune only the planning head and its selection head. Following the multimodal supervision strategy in \cite{ren2026selfsupervised}, we adopt a winner-takes-all (WTA) assignment \cite{xu2025annealed} to identify the predicted trajectory with the lowest displacement error to the refined target. The selected trajectory is optimized using a Gaussian negative log-likelihood loss, while a cross-entropy loss supervises the selection head to favor the selected mode. The distillation objective is defined as
\begin{equation}
    \mathcal{L}_{\mathrm{distill}}
    = \mathcal{L}_{\mathrm{WTA}}
    + \mathcal{L}_{\mathrm{select}}.
    \label{eq:policy_distillation}
\end{equation}
The frozen deployed model and learned trajectory scorer are used only for offline target refinement, introducing no additional inference overhead.
\begin{figure}[!hbtp]
    \centering
    \begin{subfigure}{0.45\columnwidth}
        \centering
        \includegraphics[width=\linewidth]{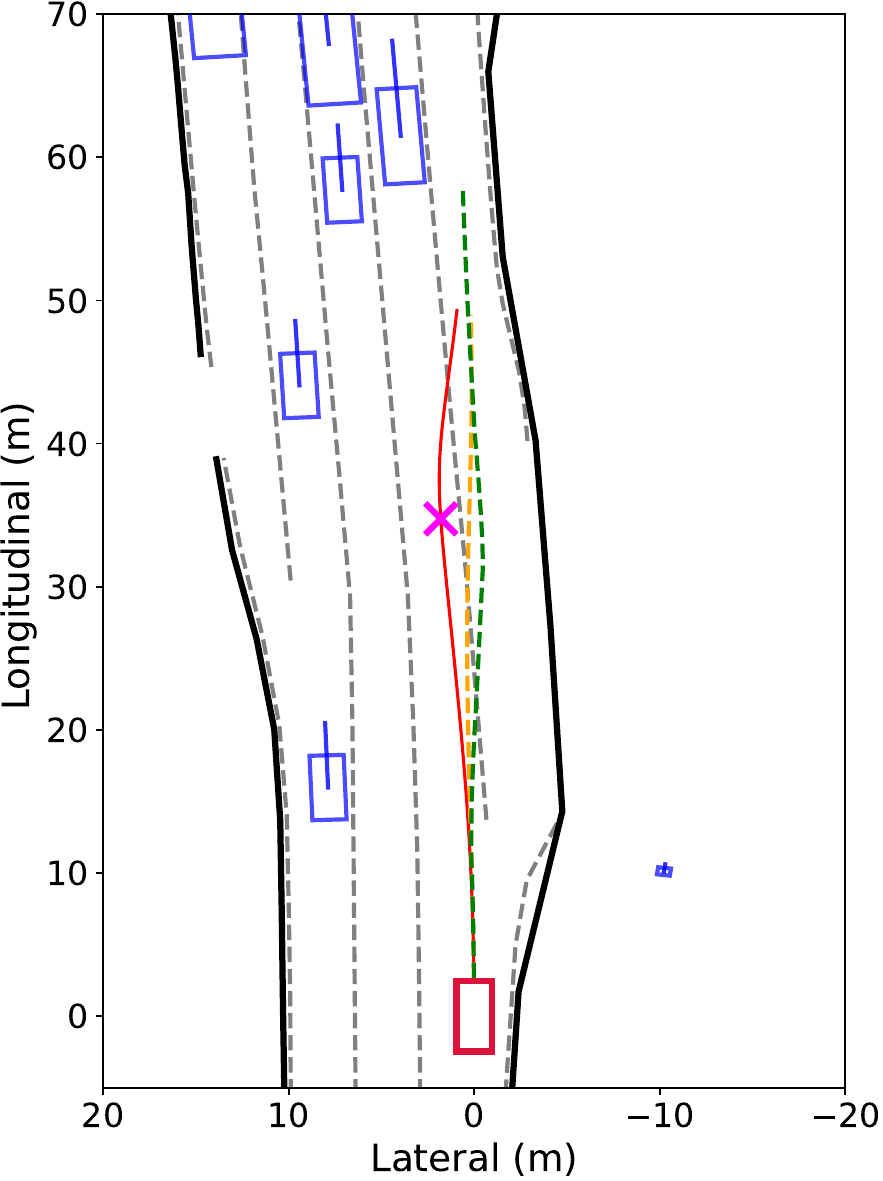}
        \caption{}
        \label{fig:target_a}
    \end{subfigure}%
    \hfill
    \begin{subfigure}{0.45\columnwidth}
        \centering
        \includegraphics[width=\linewidth]{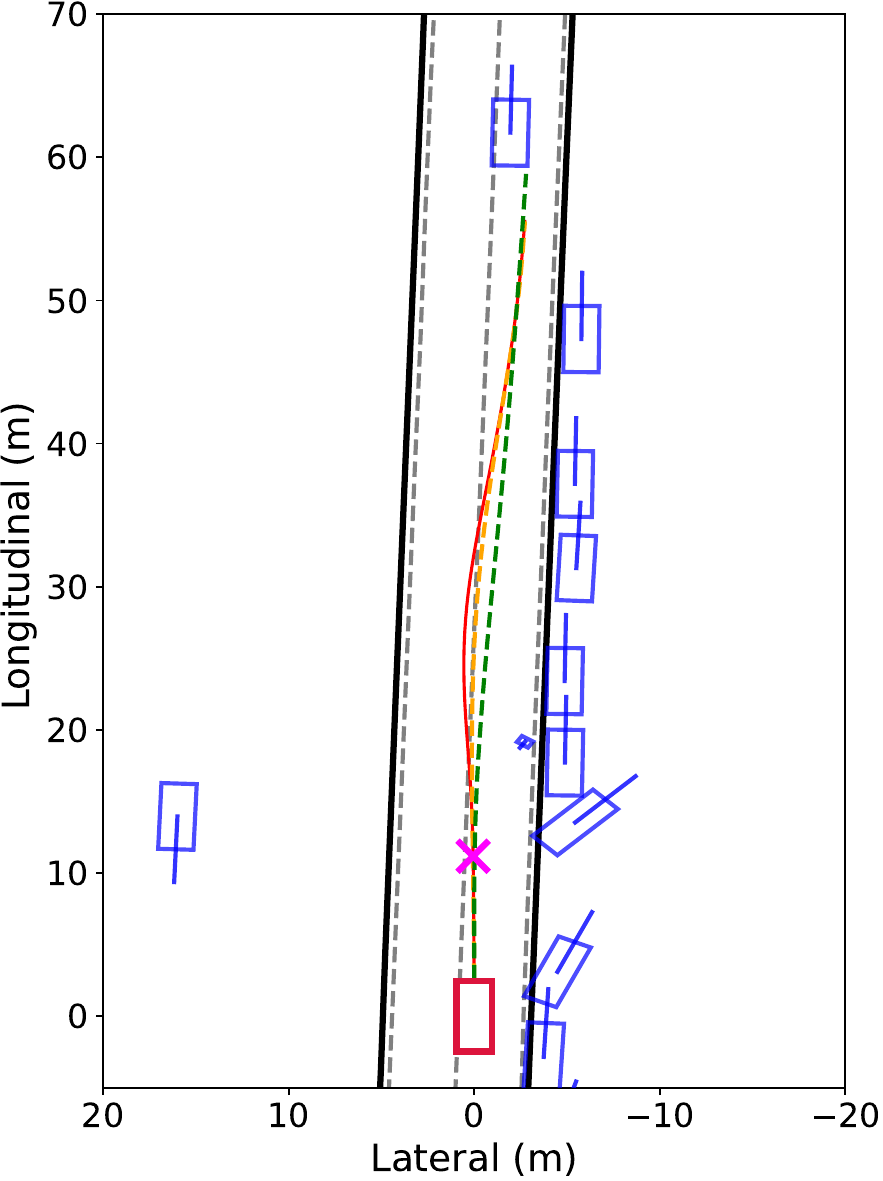}
        \caption{}
        \label{fig:target_b}
    \end{subfigure}


    \begin{subfigure}{0.45\columnwidth}
        \centering
        \includegraphics[width=\linewidth]{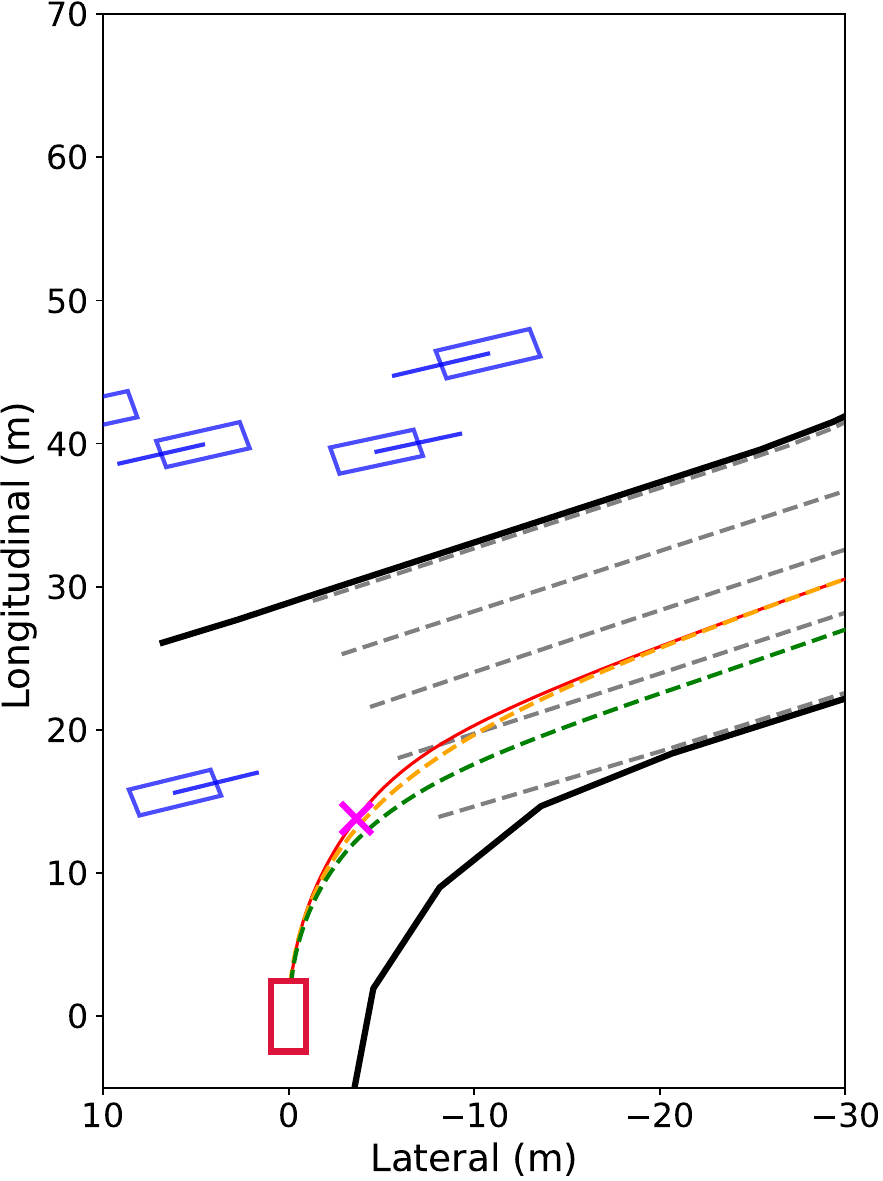}
        \caption{}
        \label{fig:target_c}
    \end{subfigure}%
    \hfill
    \begin{subfigure}{0.45\columnwidth}
        \centering
        \includegraphics[width=\linewidth]{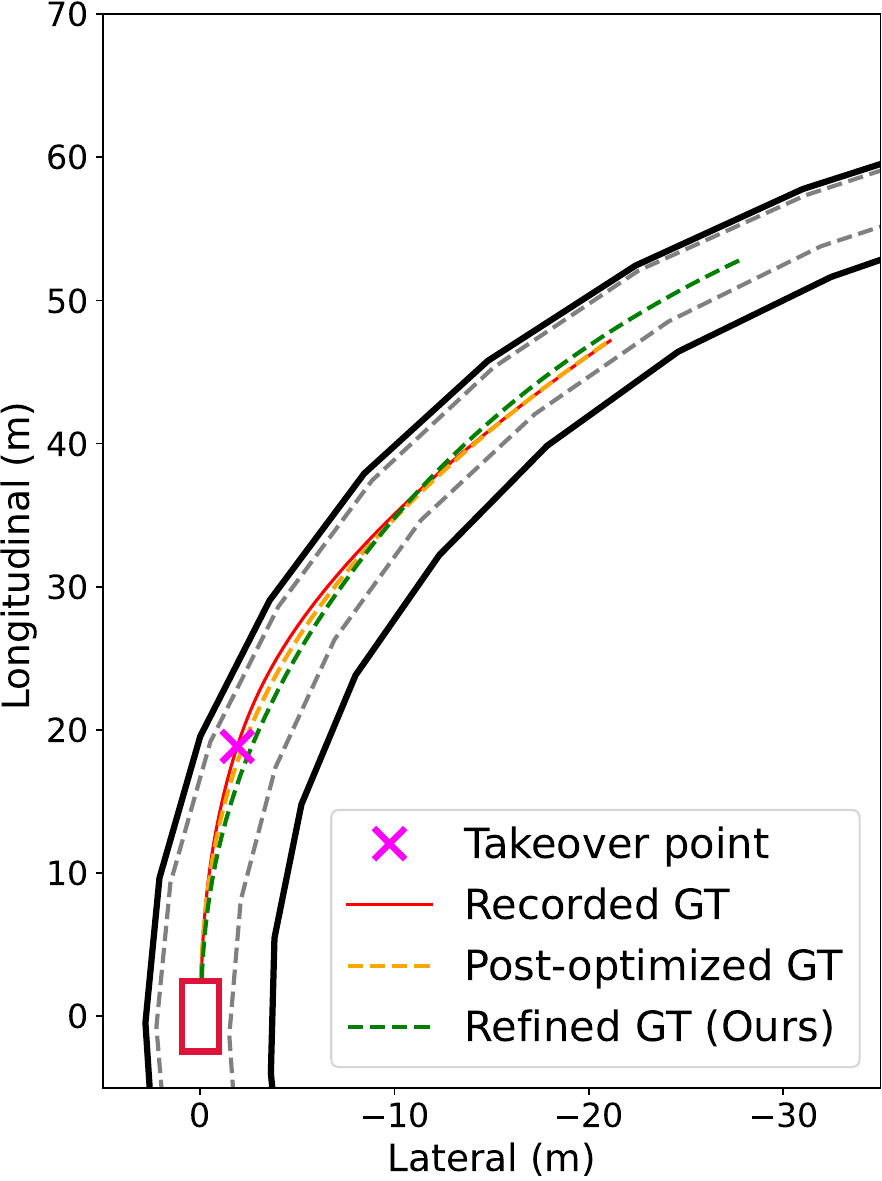}
        \caption{}
        \label{fig:target_d}
    \end{subfigure}

    \caption{Qualitative comparison of supervision targets in the four takeover scenarios shown in Figure~\ref{fig:traj_aug}. Red, green, and orange denote the recorded ground truth, our scorer-refined target, and the post-optimization target, respectively.}
    \label{fig:target_refine}
\end{figure}

Figure~\ref{fig:target_refine} compares the supervision targets in the same four takeover scenarios shown in Figure~\ref{fig:traj_aug}. Our method selects earlier corrective maneuvers for right-turn navigation and pedestrian avoidance, while favoring trajectories that better account for oncoming traffic and road boundaries on a curved ramp. Compared with conventional post-optimization, these results illustrate the advantage of scene-adaptive target refinement guided by learned trajectory desirability rather than predefined optimization objectives.

\section{Experiments}
\label{sec:experiment}

\subsection{Experimental setup}

\textbf{Evaluation settings.}
We evaluate the learned trajectory scorer through expert-aligned selection and the distilled driving model through closed-loop driving. For scorer evaluation, we use a held-out expert driving dataset without takeovers, comprising 15,202 valid frames from 154 driving segments collected in real-world urban environments. The dataset covers lane keeping (48.5\%), lane changes (33.7\%), cruising at a set speed or car following (6.1\%), intersection left turns (4.3\%), and intersection right turns (7.4\%). The candidate set excludes the original GT but includes its perturbation variants. The original GT serves as the reference for computing trajectory errors and identifying the oracle candidate.

All methods and ablations are evaluated on a shared closed-loop benchmark constructed from held-out real-world takeover logs using 4D Gaussian Splatting (4DGS) \cite{wu20244d}.
The benchmark comprises 596 challenging driving scenarios across five categories: cut-in (41.1\%), meeting (9.6\%), bypass (8.4\%), junction (32.2\%), and unprotected left turn (ULT, 8.7\%).
The driving model is deployed as a TensorRT engine on an industrial closed-loop simulation platform.
The platform updates the ego state from model plans and renders camera images and LiDAR point clouds at the updated pose, while surrounding traffic follows logged trajectories.
Runs invalidated by hardware issues are excluded, and all methods within each comparison are evaluated on the same set of valid cases.
Success rate is defined as the fraction of valid cases passing automated checks, including trajectory-output validity and collision checks against dynamic objects, static obstacles, and curbs.


\textbf{Baselines and metrics.} For trajectory scorer evaluation, we compare random selection, rule-based selection \cite{dauner2024navsim}, and learned scorers trained with different supervision objectives: CE-only, PP-only, CE+PP and Ours. Here, PP denotes paired preference supervision inspired by TakeAD \cite{liu2026takead}. We additionally report a candidate oracle that selects the trajectory with the lowest ADE@6s to characterize the quality of the candidate set. Performance is measured using Hit@1, average displacement error (ADE), and final displacement error (FDE) at 3s and 6s horizons. Hit@1 measures the proportion of selections matching the oracle candidate.

For closed-loop evaluation, we compare our method with the deployed E2E model (Base), Direct IL, Self-distillation, Post-optimization, Rule Refinement, and Reward Finetune. Self-distillation retrains the scorer without trajectory augmentation and selects the highest-scoring trajectory among the 32 original model trajectories as the supervision target. Rule Refinement selects supervision targets using predefined rules \cite{dauner2024navsim}. Post-optimization refines targets through trajectory optimization, while Reward Finetune directly optimizes the planning head by backpropagating through a frozen scorer \cite{clark2024directly}. All post-training methods use the same initial checkpoint, training data, trainable modules, and training budget.

\textbf{Implementation details.}
All experiments are conducted on 16 NVIDIA H20 GPUs. The trajectory scorer and driving model are trained for 4 and 2 epochs, respectively, with per-GPU batch sizes of 20 and 12. Both stages use the AdamW optimizer with cosine learning rate decay from $1\times10^{-5}$ to $1\times10^{-8}$. For trajectory augmentation, the frozen deployed model provides 32 trajectory proposals, with $N_{\mathrm{aug}}=3$ perturbations per proposal and $N_e=5$ human correction shifts from 0.5 to 2.5 seconds at 0.5-second intervals. The longitudinal and lateral perturbations $(\epsilon_{\mathrm{long}},
\epsilon_{\mathrm{lat}})$ are independently sampled from $\mathcal{N}(0,0.3^2)$. During distillation, only the planning head and its original selection head are finetuned, while all other model components remain frozen.

\subsection{Trajectory scorer evaluation}
\label{sec:scorer_evaluation}

As shown in Table~\ref{tab:baselines3}, all methods rank the same 133 candidates, comprising 128 model-derived trajectories and five perturbation variants of the recorded GT. The candidate oracle, which selects the trajectory with the lowest ADE@6s, achieves an ADE@6s of 0.327 m and an FDE@6s of 0.832 m, indicating that the candidate pool contains high-quality trajectories closely aligned with expert driving. In contrast, random and rule-based selection yield substantially larger errors, highlighting the importance of effective trajectory scoring.
\begin{table}[!th]
\centering
\caption{Expert-aligned trajectory selection on the expert driving set. All methods rank the same 133 candidates.}
\label{tab:baselines3}

\footnotesize
\setlength{\tabcolsep}{5.0pt}
\renewcommand{\arraystretch}{1.1}
\begin{tabular}{@{}lccccc@{}}
\toprule
Method & Hit@1 & ADE@3s & ADE@6s & FDE@3s & FDE@6s \\
\midrule
Oracle$^{\dagger}$  & 100\%  & 0.156 & 0.327 & 0.299 & 0.832  \\
\midrule
Random     & 0.86\% & 1.879 & 5.001 & 4.231 & 12.221 \\
Rule-based       & 0.32\% & 1.432 & 3.758 & 3.229 & 8.914  \\
PP-only & \textbf{12.65\%} & 0.486 & 1.734 & 1.343 & 4.722 \\
CE-only         & 9.37\% & 0.488 & 1.745 & 1.360 & 4.765 \\
CE+PP   & 8.04\% & 0.560 & 1.911 & 1.592 & 5.213 \\
Ours                & 12.12\% & \textbf{0.452} & \textbf{1.564}
                    & \textbf{1.183} & \textbf{4.305} \\
\bottomrule
\end{tabular}

\vspace{2pt}
\parbox{\columnwidth}{\footnotesize
$^{\dagger}$ Candidate Oracle selects the trajectory
with the minimum ADE@6s.
}
\end{table}

Our method achieves the lowest trajectory errors among the learned and heuristic selectors.
Compared with CE-only, CE+UP reduces ADE@6s and FDE@6s by 10.4\% and 9.7\%, respectively, whereas CE+PP yields higher errors.
PP-only achieves a slightly higher Hit@1 than ours (12.65\% versus 12.12\%) but larger displacement errors, showing that a higher oracle matching rate does not necessarily imply lower trajectory errors. These results highlight the complementary roles of CE and UP supervision in improving trajectory selection quality.

\subsection{Closed-loop driving performance}

Table~\ref{tab:baselines2} presents the closed-loop success rates across five challenging scenario
categories. Our method achieves the highest overall success rate of 81.60\%, outperforming the deployed base model (70.87\%) by 10.73 percentage points. Direct imitation learning also improves the success rate to 75.30\%, confirming the value of human takeover data for post-training. Our further improvement demonstrates the benefit of refining imperfect takeover supervision rather than directly imitating recorded trajectories. Our method also surpasses Reward Finetune (77.34\%), which updates the planning head through a frozen scorer using trajectories generated by the evolving model. Such updates may shift the trajectory distribution away from that used to train the scorer, potentially compromising reward reliability.

\begin{table}[!ht]
\centering
\caption{Closed-loop success rate across 5
challenging scenario categories. Bypass denotes obstacle avoidance maneuvers constrained by oncoming traffic.}
\label{tab:baselines2}

\footnotesize
\setlength{\tabcolsep}{2.5pt}
\renewcommand{\arraystretch}{1.12}
\resizebox{\columnwidth}{!}{
\begin{tabular}{@{}lcccccc@{}}
\toprule
Method & Cut-in & Meeting & Bypass & Junction & ULT & Overall$^{\dagger}$ \\
\midrule
Base                & 86.94 & 61.40 & 40.82 & 67.72 & 42.55 & 70.87 \\
\midrule
Direct IL           & 84.90 & 78.95 & 38.78 & 74.07 & 63.83 & 75.30 \\
Self-distillation   & 84.90 & 73.68 & 48.98 & 68.78 & 57.45 & 73.42 \\
Post-optimization  & 81.63 & \textbf{84.21} & 40.82 & 70.90 & 42.55 & 71.89 \\
Rule Refinement     & 86.12 & 75.44 & 48.98 & 71.43 & 59.57 & 75.13 \\
Reward Finetune           & 88.16 & 73.68 & 36.73 & 76.72 & 70.21 & 77.34 \\
\midrule
Ours                & \textbf{92.65} & 82.46 & \textbf{51.02}
                    & \textbf{77.25} & \textbf{72.34} & \textbf{81.60} \\
\bottomrule
\end{tabular}
}
\vspace{2pt}
\parbox{\columnwidth}{\footnotesize
$^{\dagger}$All methods are evaluated on the same 587 valid cases.
}
\end{table}


Our method achieves the best performance in four of the five scenario categories, with particularly substantial improvements in unprotected left turns and meeting scenarios. It also achieves the highest success rate in challenging bypass scenarios involving obstacle avoidance under oncoming traffic. Notably, post-optimization performs slightly better in meeting scenarios, where relatively predictable oncoming motion favors predefined collision-avoidance objectives. However, its limited improvements in other scenarios suggest that predefined optimization objectives struggle to accommodate diverse driving interactions. In contrast, our method achieves more consistent improvements through scorer-guided refinement of takeover supervision.

\begin{figure*}[!th]
  \centering
  \captionsetup[subfigure]{
    font=footnotesize,
    labelformat=parens,
    labelsep=space,
    justification=centering,
    singlelinecheck=false,
    skip=3pt
  }


  \begin{minipage}[c]{0.26\linewidth}
    \centering
    \begin{subfigure}{\linewidth}
      \centering
      \includegraphics[width=\linewidth, trim=0 85mm 0 5mm, clip]{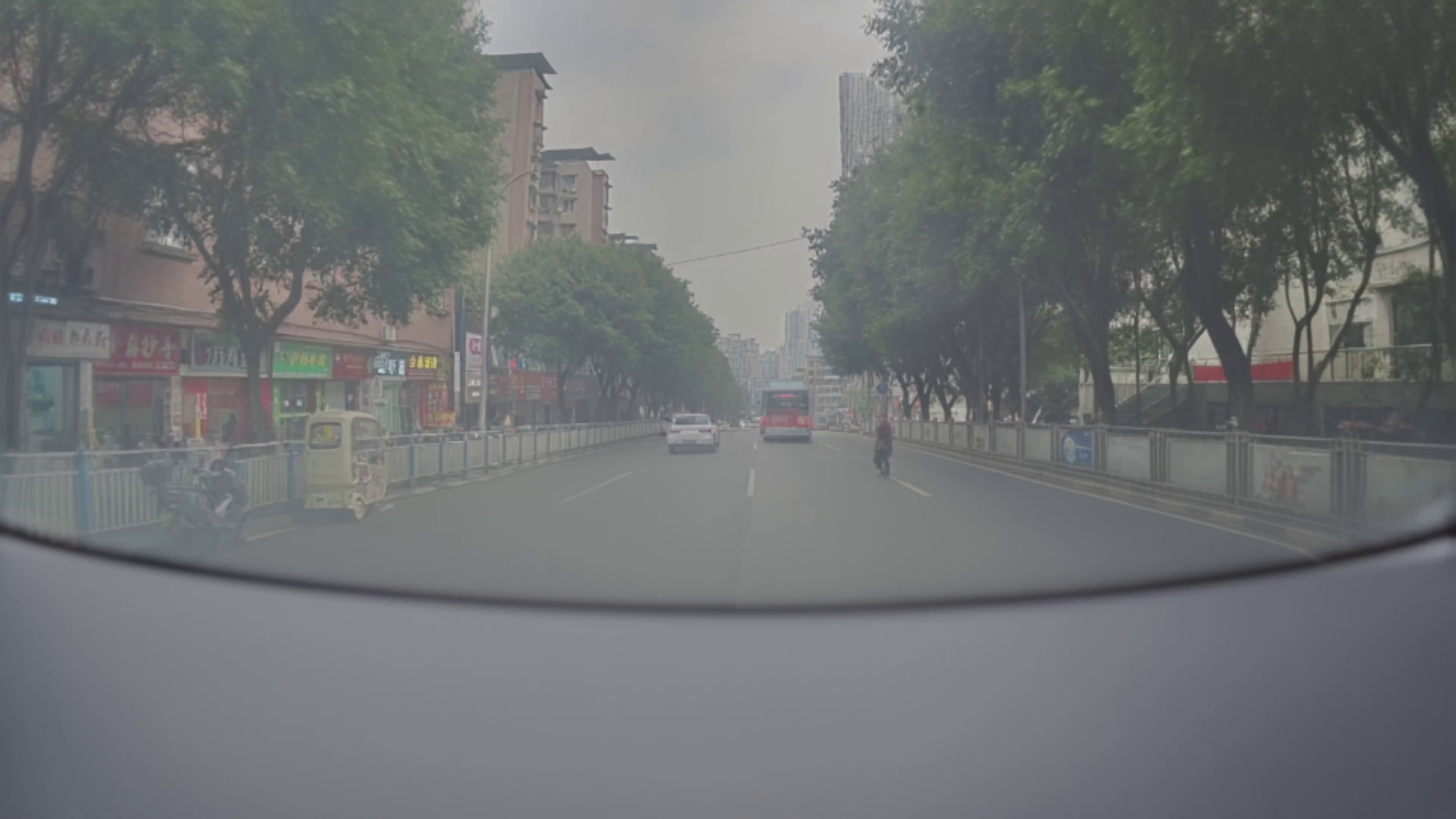}
      \caption{$T_1$: Pedestrian intrusion onset.}
      \label{fig:base-t0}
    \end{subfigure}

    \par\vspace{0pt}

    \begin{subfigure}{\linewidth}
      \centering
      \includegraphics[width=\linewidth, trim=0 85mm 0 5mm, clip]{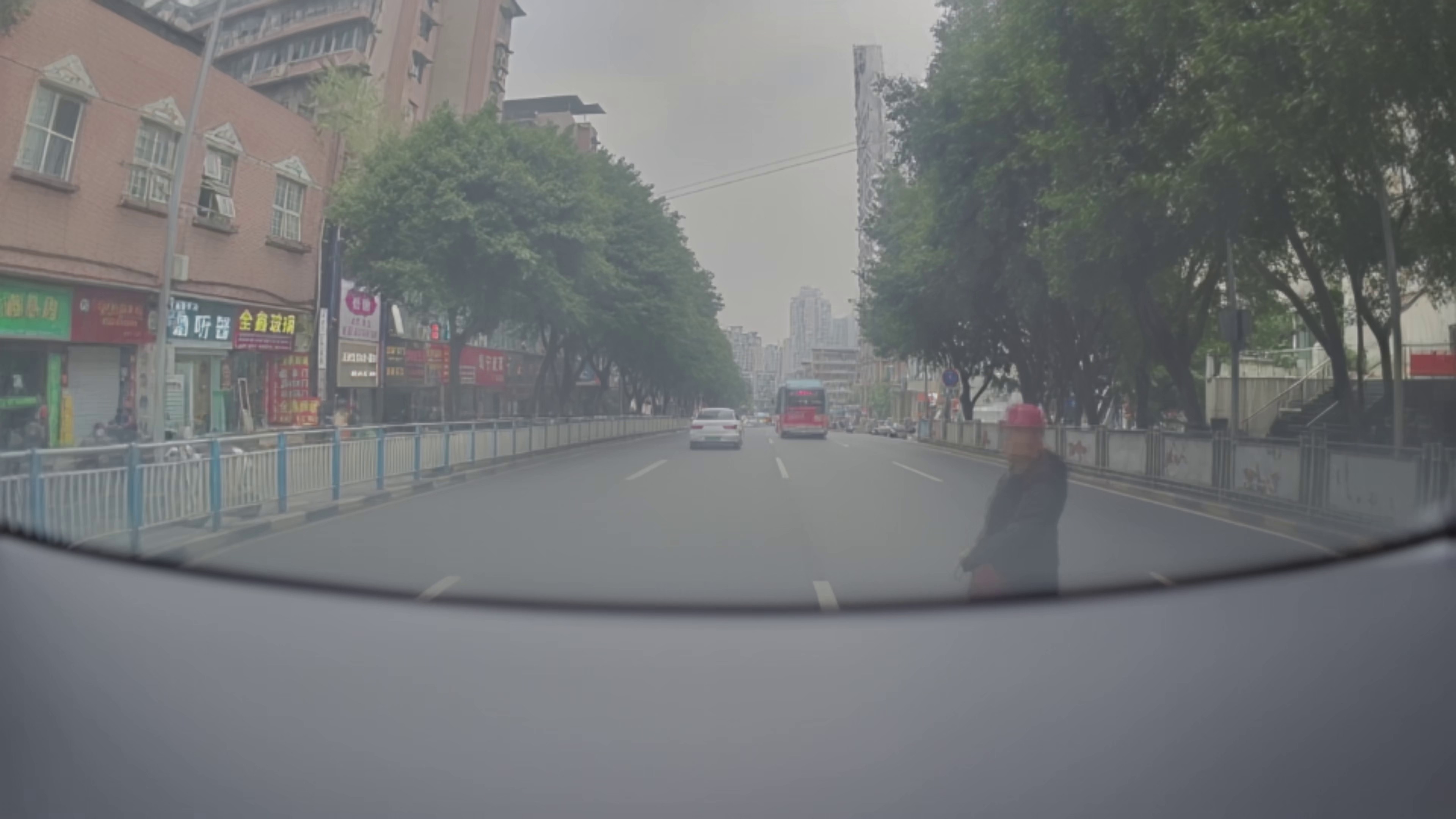}
      \caption{$T_2$: Collision.}
      \label{fig:base-t1}
    \end{subfigure}
  \end{minipage}%
  \hspace{0.0\linewidth}%
  \begin{subfigure}[c]{0.71\linewidth}
    \centering
    \includegraphics[width=\linewidth, trim=0 0mm 0 2mm, clip]{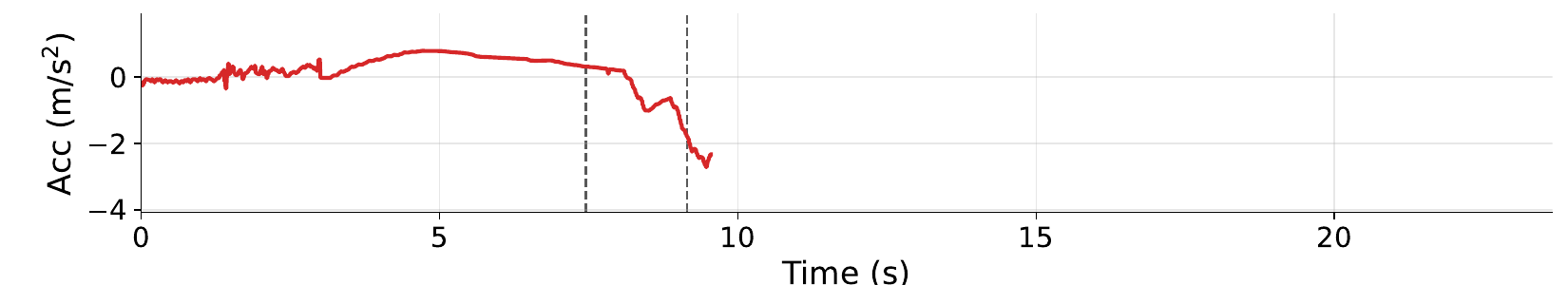}
    \par\vspace{0pt}
    \includegraphics[width=\linewidth, trim=0 0mm 0 2mm, clip]{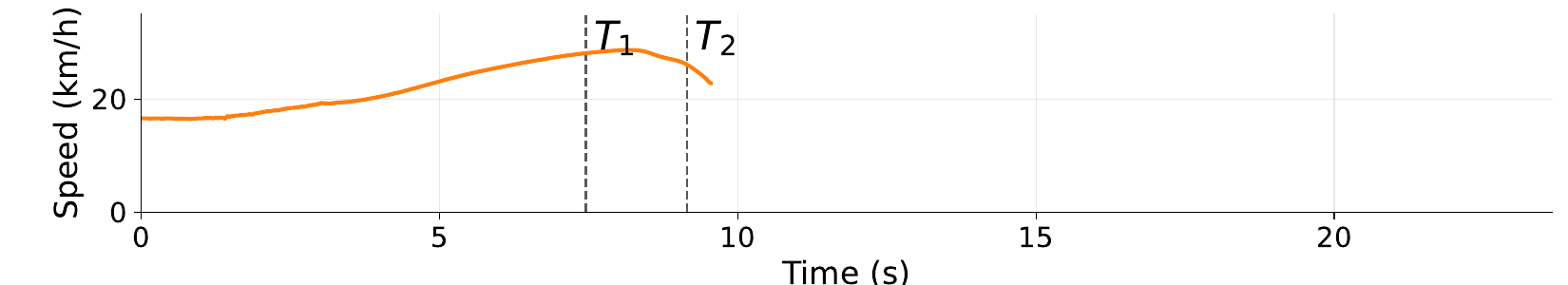}
    \caption{Acceleration and speed of the base model.}
    \label{fig:base-curves}
  \end{subfigure}

  \par\vspace{0pt}


  \noindent
  \begin{minipage}[c]{0.26\linewidth}
    \centering
    \begin{subfigure}{\linewidth}
      \centering
      \includegraphics[width=1.0\linewidth, trim=0 85mm 0 5mm, clip]{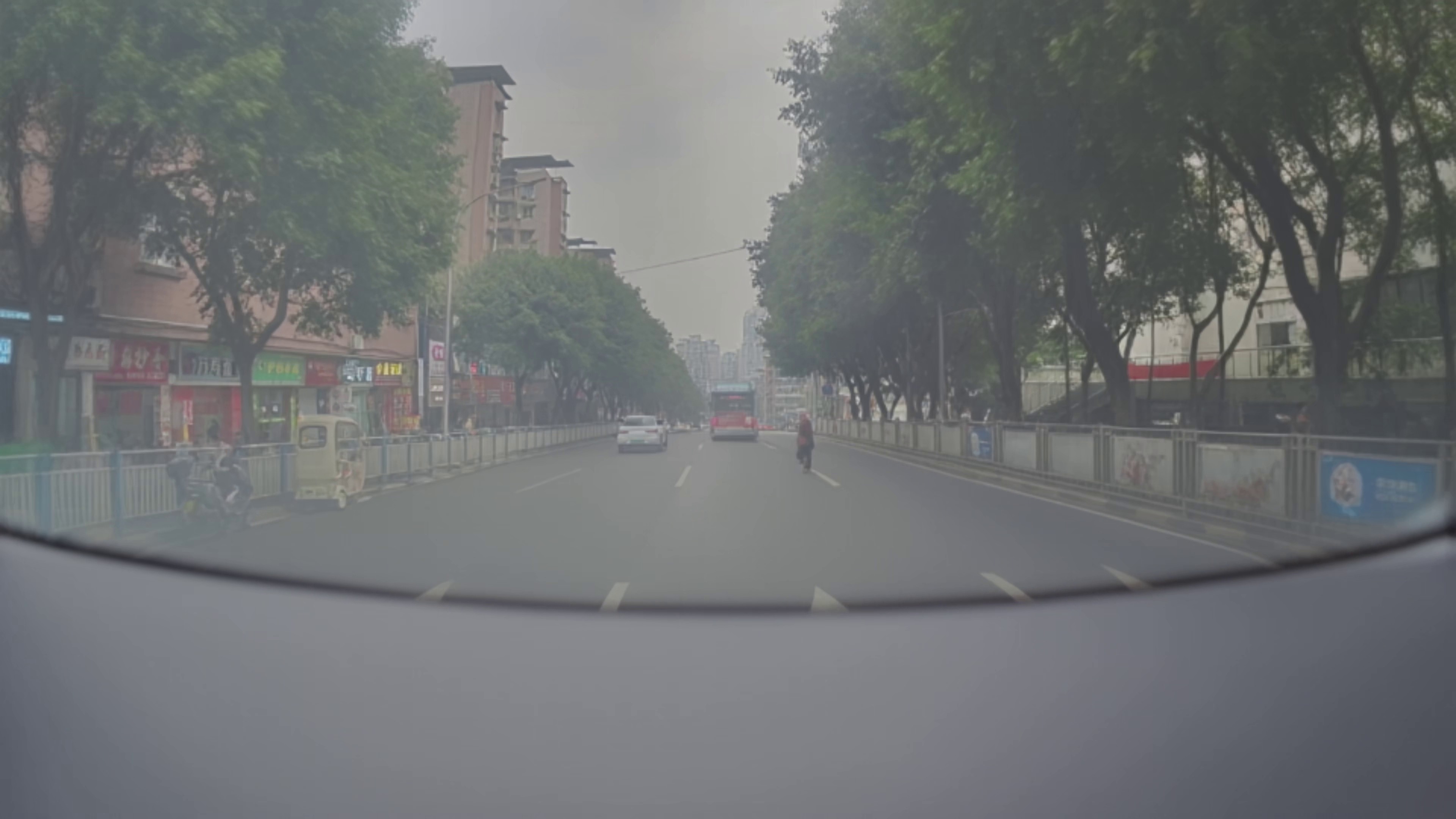}
      \caption{$T_1$: Pedestrian intrusion onset.}
      \label{fig:we-t0}
    \end{subfigure}

    \par\vspace{0pt}

    \begin{subfigure}{\linewidth}
      \centering
      \includegraphics[width=1.0\linewidth, trim=0 85mm 0 5mm, clip]{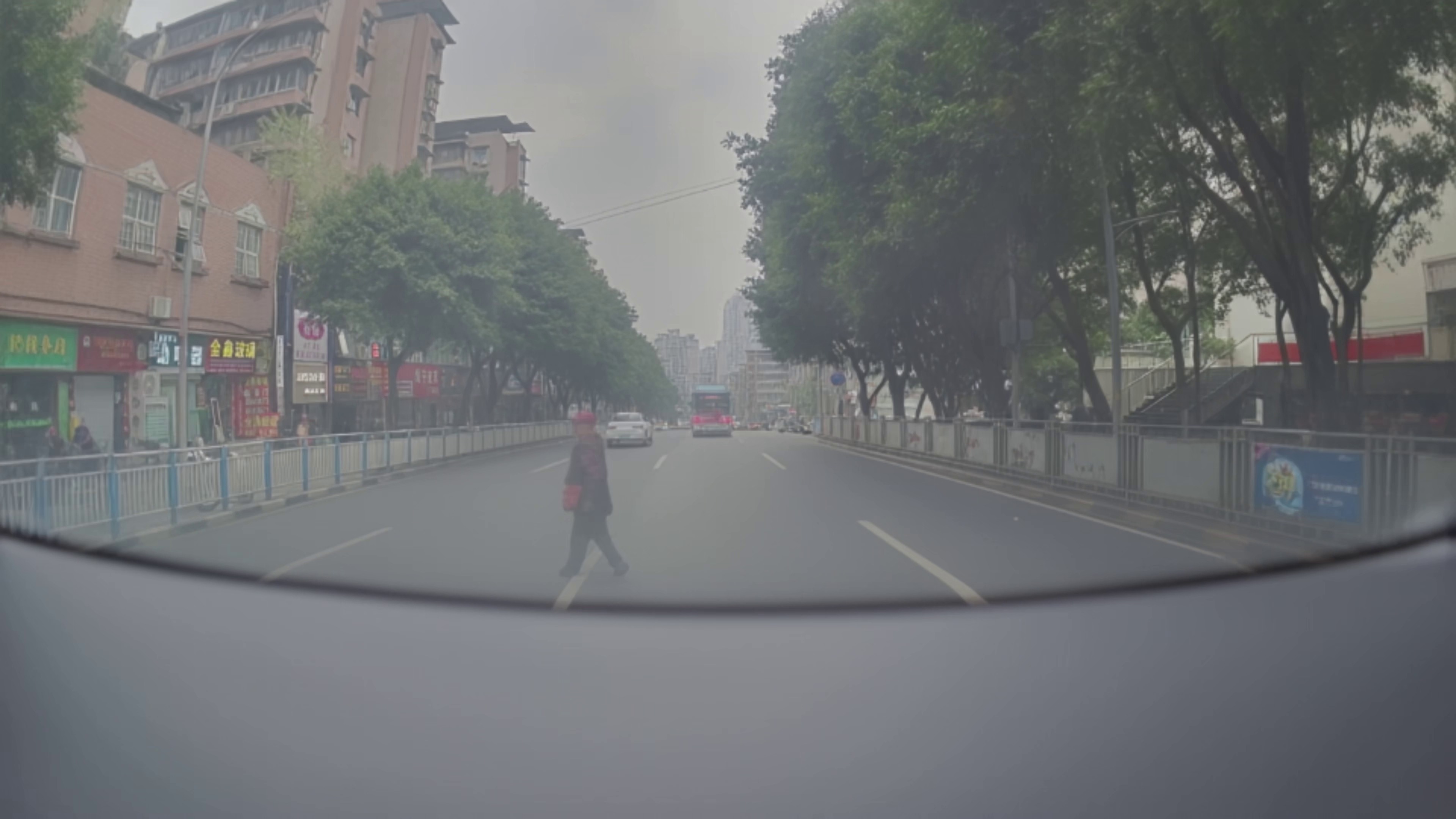}      
      \caption{$T_2$: Safe yielding.}
      \label{fig:we-t1}
    \end{subfigure}
  \end{minipage}%
  \hspace{0.0\linewidth}%
  \begin{subfigure}[c]{0.71\linewidth}
    \centering
      \includegraphics[width=1.0\linewidth, trim=0 0mm 0 2mm, clip]{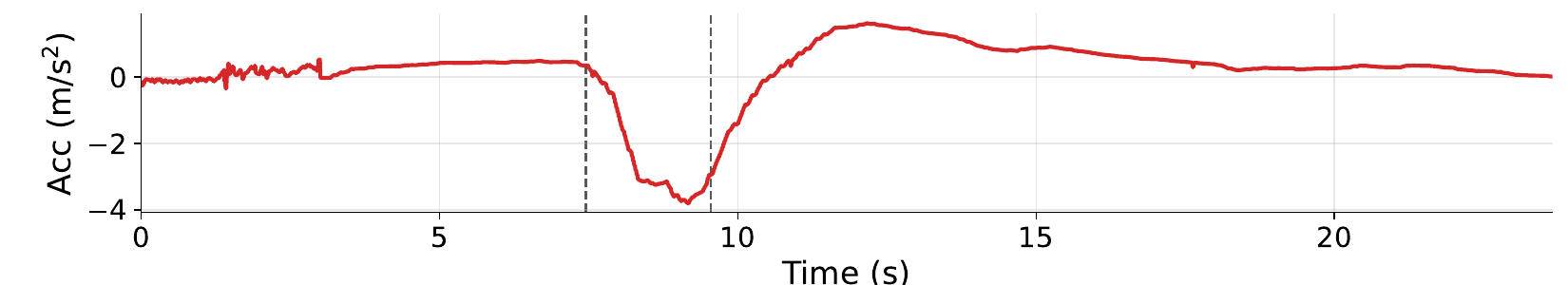}    
      \par\vspace{0pt}
      \includegraphics[width=1.0\linewidth, trim=0 0mm 0 2mm, clip]{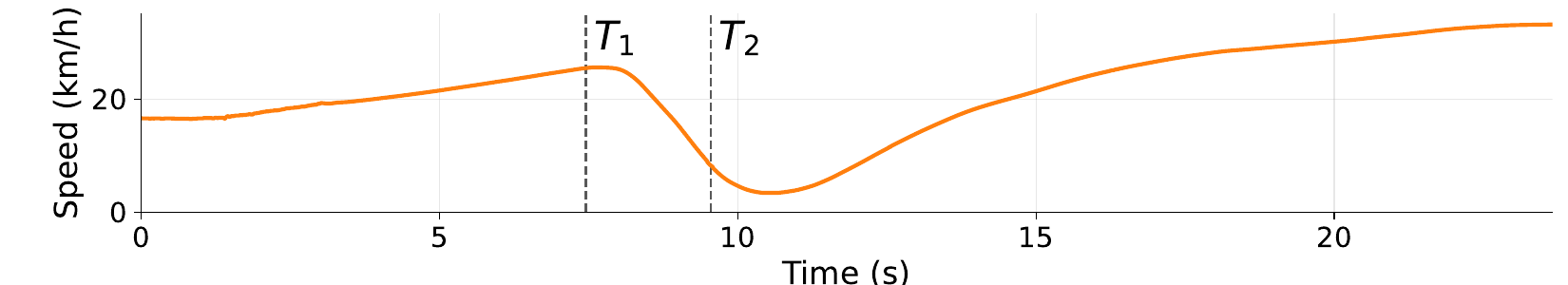}    
      \caption{Acceleration and speed of our model.}
    \label{fig:we-curves}
  \end{subfigure}

    \caption{Qualitative comparison in a 4DGS-based closed-loop pedestrian avoidance scenario. The top and bottom groups show the deployed base model and our post-trained model, respectively. The left panels show two key camera frames ($T_1$ and $T_2$) selected separately for each model, with the corresponding acceleration and vehicle speed on the right.
    }
    \label{fig:closedloop-comparison}
\end{figure*}

Figure~\ref{fig:closedloop-comparison} illustrates a representative pedestrian avoidance scenario from the closed-loop evaluation. As a pedestrian enters the driving path, the base model responds with insufficient deceleration, reaching only about $-2.5\,\mathrm{m/s^2}$ and failing to avoid a collision. In contrast, our post-trained model initiates braking earlier and more decisively, reaching approximately $-3.8\,\mathrm{m/s^2}$ and reducing its speed to nearly zero to safely yield to the pedestrian before resuming normal driving. This comparison demonstrates that post-training enables earlier and more proactive responses to potential collisions.

\subsection{Ablation studies}

\begin{table}[!ht]
\centering
\caption{Effect of trajectory augmentation on the overall closed-loop success rate. All success rates are calculated over 594 valid scenarios.}
\label{tab:augmentation_ablation}
\footnotesize
\setlength{\tabcolsep}{6pt}
\renewcommand{\arraystretch}{1.0}
\begin{tabular*}{\columnwidth}{@{\extracolsep{\fill}}lccc@{}}
\toprule
Setting & $N_{\mathrm{aug}}$ & $N_e$ & Overall(\%) \\
\midrule
Random augmentation & 0 & 5 & 73.23 \\
                    & 5 & 5 & 78.62 \\
\midrule
Correction shifting & 3 & 1 & 75.25 \\
                    & 3 & 3 & 75.93 \\
\midrule
Full (Ours)         & 3 & 5 & \textbf{81.48} \\
\bottomrule
\end{tabular*}
\end{table}

\noindent\textbf{Trajectory augmentation.}
Table~\ref{tab:augmentation_ablation} examines the effect of trajectory augmentation.
Removing random augmentation reduces the success rate from 81.48\% to 73.23\%, while increasing $N_{\mathrm{aug}}$ from 3 to 5 lowers it to 78.62\%. This non-monotonic trend shows that enlarging the candidate pool alone does not guarantee better closed-loop performance.
With $N_{\mathrm{aug}}=3$, increasing $N_e$ from 1 to 3 and 5 improves the success rate from 75.25\% to 75.93\% and 81.48\%, respectively.
More early correction variants cover a wider range of temporal offsets, allowing the scorer to choose among different correction timings for each scenario.
These results favor a moderate number of random perturbations and multiple early correction variants.

\noindent\textbf{Scorer supervision and target refinement.}
Table~\ref{tab:scorer_ablation} compares different scorer supervision objectives and target replacement strategies. GT replacement rates are measured over ten training batches, considering samples within 6 seconds before takeover. CE+PP yields the lowest replacement rate because paired preference supervision treats recorded takeover trajectories as positive samples. CE-only lacks explicit desirability supervision, resulting in more frequent replacements (39.81\%) but limited closed-loop performance (77.16\%). To examine selective refinement, Ours+UR removes the temporal window and score-margin constraints in Equation~\ref{eq:target_refinement}, replacing GT whenever a candidate scores higher during autonomous driving. Despite its much higher replacement rate (77.87\%), Ours+UR achieves a lower success rate (79.36\%) than ours (81.39\%), which replaces only 34.39\% of targets. These results highlight the importance of both unpaired preference supervision and selective target refinement.

\begin{table}[!ht]
\centering
\caption{Effect of scorer supervision on target refinement and closed-loop success rate. All settings are evaluated on the same 591 valid cases.}
\label{tab:scorer_ablation}
\footnotesize
\setlength{\tabcolsep}{4pt}
\renewcommand{\arraystretch}{1.0}
\begin{tabular*}{\columnwidth}{@{\extracolsep{\fill}}lcccc@{}}
\toprule
Setting & CE+PP & CE-only & Ours+UR & Ours\\
\midrule
Success rate (\%)              & 76.31 & 77.16 & 79.36 & 81.39 \\
GT replacement rate (\%)       & 28.98 & 39.81 & 77.87 & 34.39 \\
\bottomrule
\end{tabular*}
\end{table}

\section{Conclusion}
\label{sec:conclusion}

In this paper, we proposed an offline post-training framework that transforms imperfect real-world human corrections into proactive supervision for E2E driving models. By integrating intervention-aware trajectory augmentation with a scene-conditioned preference scorer, our method selectively refines pre-takeover supervision targets and distills them into the driving model without additional inference overhead. Closed-loop evaluations in challenging scenarios demonstrate substantial improvements over the deployed model, direct imitation learning, and trajectory post-optimization.
Importantly, our framework directly exploits existing fleet logs without requiring additional manual trajectory annotations or sensor-level scene reconstruction for training. This offers a practical path toward automated and efficient model updates that can be delivered to production vehicles through over-the-air updates. Future work will extend post-training to the entire E2E model and incorporate scenario generation techniques to address extremely rare corner cases. 
\newpage

{
    \small
    \bibliographystyle{ieeenat_fullname}
    \bibliography{main}
}


\end{document}